\documentclass{IEEEtran}

\usepackage{cite}

\ifCLASSINFOpdf
  \usepackage[pdftex]{graphicx}
  \DeclareGraphicsExtensions{.pdf,.jpg,.png}
\else
\fi
\usepackage[cmex10]{amsmath}
\usepackage{amsfonts}
\usepackage{algorithmic}
\usepackage{algorithm}
\usepackage{multirow}

\begin{document}

\title{A Parameter-free Affinity Based Clustering}

\author{
        Bhaskar Mukhoty, Ruchir Gupta,~\IEEEmembership{Member IEEE} and Y. N. Singh,~\IEEEmembership{Senior Member IEEE}
%
\\The paper is under consideration at Pattern Recognition Letters.

}

\maketitle
\begin{abstract}
Several methods have been proposed to estimate the number of clusters in a dataset; the basic ideal behind all of them has been to study an index that measures inter-cluster separation and intra-cluster cohesion over a range of cluster numbers and report the number which gives an optimum value of the index. In this paper we propose a simple, parameter free approach that is like human cognition to form clusters, where closely lying points are easily identified to form a cluster and total number of clusters are revealed. To identify closely lying points, affinity of two points is defined as a function of distance and a threshold affinity is identified, above which two points in a dataset are likely to be in the same cluster. Well separated clusters are identified even in the presence of outliers, whereas for not so well separated dataset, final number of clusters are estimated and the detected clusters are merged to produce the final clusters. Experiments performed with several large dimensional synthetic and real datasets show good results with robustness to noise and density variation within dataset.
\end{abstract}




%


\section{Introduction}

Cluster analysis is an unsupervised learning problem, where the objective is to suitably group $n$ data points, $X=\{x_1,x_2,...,x_n\}$, where $ x_i$ is taken from a d-dimensional real space $\mathbb{R}^d$. Traditionally this has been considered as an important statistical problem with numerous applications in various fields including image analysis, bio-informatics and market research.\\
  Generally, if the number of clusters $k$ is given, finding the optimal set of  clusters 
$C=\{c_1,c_2,...c_k\}$ which minimizes the variance or sum of squares of distances from the points to the center of their clusters, is an NP-hard problem\cite{nphard}. Here points represent some parametric measure for each data point represented in the d-dimensional real space. The variance or sum of squares within ($SSW$) is given as
\begin{equation}
 SSW(C)= \sum\limits_{j=1}^k \sum\limits_{x_i \in c_j}||x_i - \bar{c_j}||^2
\end{equation}

 Here, $||p|| $ denotes the magnitude of a vector $p$, so that $||p-q||$ is the euclidean distance between two points $p$ and $q$. The point $\bar c_j$ denotes mean of $x_i$ s.t. $x_i \in c_j$, or the centroid of the cluster $c_j$. It can be observed that if SSW is calculated by varying total number of clusters, it reduces with the increase of number of clusters, and ultimately goes to zero, when the number of clusters becomes equal to number of data points $n$.
Methods like Gap Statistics\cite{gap}, study the curve of SSW plotted against number of cluster, for an elbow point after which the curve flattens, to give an optimal number of clusters.

Internal cluster evaluation indexes like Silhoutte\cite{silhouette}, Calinsky-Harabasz \cite{CH} evaluate quality of clustering depending upon some measures of intra-cluster cohesion and inter-cluster separation. They are often used to predict suitable number of clusters by searching the data with a range of cluster numbers and reporting the number which gives optimal value of the index\cite{cvi_comp}. This, when followed by a clustering algorithm, parameterized by number of clusters, gives partition of the data\cite{kmeans}.


Other parameterized algorithms supplied with some form of information about the data, are also in use. DBSCAN\cite{dbscan}, a density based clustering algorithm requires parameters like neighborhood radius and minimum number of points in that radius to identify core point of the clusters. Generally, additional work is needed to estimate such parameters which requires area specific domain knowledge\cite{apscan}.

In order to have a parameter free clustering algorithm, we have taken a different approach to identify the clusters. Our method imitates the way human recognizes the groups in the data. Human, when exposed to a representation of an intelligible dataset, at once recognizes the clusters present, because some data points appears so close, that they could hardly go to different clusters. Such groups when counted give the number of clusters. We do not take the redundant approach to search through space of possible clusterings, to identify optimal clusters.

Our algorithm tries to imitate human cognition. In order to identify closely grouped points we calculate an affinity threshold followed by sequential search within the data space for the points in vicinity, which leads to the formation of groups. Points, which remain single in such a search, are identified as outliers. If the data has well separated clusters, the said process will be able to detect them, whereas, if the clusters are close enough they could also be merged to form new cluster. The nature of the dataset is decided by cost function defined in section 3.2. Unlike equation 1, the cost will not always increase with the decrease in number of clusters, indicating that the dataset supports merge. For such datasets we prioritize large clusters as representative of the data and the number of final cluster is estimated using distribution of detected cluster sizes. Clusters are then merged in order of closeness to produce final clusters.



We have conducted experiments with standard convex datasets, and compared the performance of the algorithm with the existing algorithms, in terms of proposed number of clusters and quality of the obtained clusters.

The remainder of the paper is organized as follows. Section 2 introduces the previous works related to clustering algorithms. Section 3 describes the proposed parameter free algorithm. Section 4 experimentally evaluates the relative performance of the proposed algorithm, Section 5 presents the conclusions.

\section{Related Work}
The general problem of grouping of data can be translated to different mathematical interpretations. For some datasets that do not have well separated clusters the reported  clusters may differ depending upon the way the problem is defined. Following are the four basic kind of approach taken to address cluster analysis problem.
\subsection{Partition Method}
This kind of methods partition the data into k Voronoi cells\cite{Voronoi}, where k is supposed to be known. Starting with k arbitrary cluster centers k-means allocates points to the cluster which has nearest center. After allocation if the centroid of a cluster, shifts from the previous cluster center, the centroid becomes the new cluster center and points are reallocated. The last step repeats till the process converges\cite{kmeans,kmeans++}.\\
Although k-means and its variations are quite fast, these methods require number of clusters as parameter. Moreover they are not deterministic, that is, these may produce different results on different runs of the algorithm, so aggregation of results often becomes necessary\cite{concensus}. Out-lier detection is also not possible in this format.

\subsection{Hierarchical Method}
In these methods, the clusters are formed either by agglomerating nearest points in a bottom up approach or by partitioning the set in a divisive top down approach, until it gives the desired number of clusters or falsify some parametric criterion which measures separation and cohesion of clusters\cite{chameleon}. The agglomeration or division typically depends upon the distance measure used to define separation between the clusters. Among the several distance measure available, single-linkage defines the cluster distance as the minimum distance between all pair of points taken from the two clusters\cite{slink}, whereas complete-linkage takes the maximum. Average-linkage defines the cluster distance as the sum of all pairs of distance normalized by the product of the cluster sizes\cite{avg}. Ward method measures the increase in SSW when two clusters are merged\cite{ward}.

Both agglomerative and divisive approach requires a stopping criteria to produce required number of clusters. The criteria may be the number of cluster itself or some internal cluster validity index(CVI). Computing the CVI at every step of merge of the clusters are costly. Further these methods are not robust to outliers.

\subsection{Distribution Method}
In this kind of method data points are assumed to come from a certain distribution. In Gaussian Mixture Model, clusters are assumed to come from Gaussian distributions, and expectation maximization algorithm\cite{em} parameterized with number of clusters is used to estimate distributions parameters. Random outliers can severely degrade performance of such algorithm by giving misplaced cluster centers.

\subsection{Density Method}
Density based methods such as DBSCAN\cite{dbscan} method works by detecting points in a cluster using a neighborhood radius length $\epsilon$,  and  a parameter $minPts$ to represent minimum number of points within the neighborhood. This method although robust to outliers and works for arbitrary shapes, is susceptible to density variations. OPTICS\cite{optics} is an improvement which overcomes the requirement of $\epsilon$ as parameter. Finding suitable value for density parameter in such algorithm requires domain knowledge.

Our method, in contrast being parameter free does not require any additional input from the user. Moreover by the nature of cluster identification the method gives immunity to noise and identify the outliers present in the dataset. All this happens in a time bound manner which is no more than the time required by most hierarchical algorithms that is of the order of $O(n^2)$.

\section{Method Proposed}
In the proposed algorithm we try find the natural clusters in a dataset. We define a notion for closeness in which we derive how close two points must be, to be in the same cluster. The data space is then searched to find points that are close enough to be in the same cluster. Often clusters so detected are close enough so that additional merge is required to club them. The whole task is performed in two parts, $findClusters$ method listed in Algorithm \ref{alg1} detects the initial clusters and the $mergeClusters$ method listed in Algorithm 4 further merges these clusters if required. Following two sections describe the process in detail.
\subsection{Finding initial clusters}

Before we do actual processing, feature scaling is often a necessary step to start with. In the proposed algorithm we assume numeric no categorical data as input, but the range of each dimension of such data may vary widely. Hence, while using a distance function to measure separation between two points, the dimension with higher range would dominate the function. Thus as a preprocessing each dimension of the input matrix $X_{n\times d}$, is normalized to give the standard z-score of the values, we denote the normalized input matrix by $Z_{n \times d}$.
\begin{equation}
Z^{(j)}=\dfrac{X^{(j)}-\mu_j}{\sigma_j} \qquad \forall j \in \{1..d\}
\end{equation}

 where $\mu_j$, $\sigma_j$ respectively denotes  the mean and standard deviation of $j^{th}$ column of X.

 We form a $n\times n$ distance matrix D to store euclidean distance between all pair of points from normalized data matrix Z. 
 
\begin{equation}
	D(i,j)=||Z_i-Z_j||  
\end{equation}

where $Z_i$ denotes $i^{th}$ row of the matrix Z.

We would now define an affinity function A, so as the pair of nodes which are closer will have an affinity near one, and nodes which are farther will have an affinity near zero between them. In order to do so we transform the distance through a Gaussian function to get a half bell shaped curve which slowly approaches zero as distance goes to infinity. The bell curve would ensure distances near zero are having higher affinity. Thus, the affinity is:
 
\begin{equation}
A(i,j)=exp(\dfrac{-D(i,j)^2}{2\sigma_D'^2})
\end{equation}	
	
	$\sigma_D'$ is square root of standard deviation $\sigma_D$, of the elements of D. In the distance matrix D, there exists not only intra-cluster distances for which Gaussian distribution is reasonable but also the inter-cluster distances. This leads to higher standard deviation for a cluster. Thus by taking $\sigma_D'=\sqrt{\sigma_D}$ in the affinity function, we ensure a relatively flattened bell curve. 
	

 We then take a histogram H of all affinity scores, which divides the score in $b$ bins of $1/b$ range. Bin analysis performed in section 4.4 shows that  taking number of bins b as 10 yields best results.
 
We observed that among affinities between all pair of points, there has to be a high frequency of strong affinities as points within the same cluster would have affinity near 1. But how prominently they are near to unity will depend upon how condensed the clusters are. Thus we seek for an affinity threshold where maximum positive change in affinity count occurs. Above this point affinity would be considered strong. The center of the bin (say $k$) after which the change occurs is declared as threshold, given by, \\
	   	 \begin{equation}
	   	 	threshold=((k-1)+ 1/2)/b 
	   	 \end{equation}	

 where $ k = \smash{\displaystyle\max_{ i }} (H(i+1)-H(i)), i \in \{1,..,b-1\}$. $H(i)$ is the magnitude of the histogram for bin $i$.
 
 If two points have affinity above the threshold, the points are likely to belong to the same cluster. Figure 1 shows affinity histogram taken for d8c8N\cite{noise_data} data set. The $threshold$ comes at 0.85, representing the affinity of the ninth bin, after which we can observe the increase in affinity count at tenth bin.
 	 
 	\begin{figure}
  		 \includegraphics[scale=0.45]{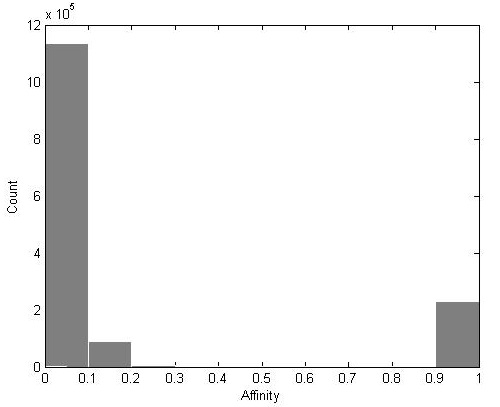}
  		\caption{Affinity histogram of d8c8N}
  		\label{fig1}
  	\end{figure}
  
  With the threshold in hand we start with an unclustered point, we check for all other points which have affinity above the threshold and include them in the cluster. Affinity of a unclustered point is calculated w.r.t the centroid of the forming clusters which is incrementally updated every time when a point is added.

  Centroid of $j^{th}$ cluster $Cnt(j)$, is conventionally defined as the mean of the points in the cluster. Thus,
  \begin{equation}
   Cnt(j)= \frac{1}{s_j}\sum\limits_{\forall i|C(i)=j}{Z_i} 
   \end{equation}
     where the vector $C(i)$ denotes the cluster Id of the point $i$ and is initialized to zero to denote that the points initially do not belong to any cluster. $s_j$ is the size of cluster $j$.
     
      Process of cluster detection continues untill all points are assigned to some cluster.  A point may shift from one cluster to other in the process, if it is closer to the  centroid of the other cluster. This shifting helps to form more cohesive clusters.
   
  Result of the above process could be to have some single point as clusters, we would identify them as outliers and separate them from future processing. Algorithm 1, $findClusters$ gives the pseudo code of the process described so far.
  
  \begin{algorithm}
  	
  	\caption{$findClusters$}
  	\label{alg1}
  	\begin{algorithmic}[1]
  		\renewcommand{\algorithmicrequire}{\textbf{Input:}}
  		\renewcommand{\algorithmicensure}{\textbf{Output:}}
  		\REQUIRE $X_{n \times d}$ (n data points in d dimensions)\\
  		\ENSURE $C_{n \times 1}$ (vector giving cluster id of each point)\\

  		\COMMENT {Normalization of $X$; $X^{(j)}$  denotes $j^{th}$ column $X$} 
  		
  		\FOR{$j=1$ to $d$} 	 
  		\STATE $\mu_j = mean(X^{(j)}); \quad \sigma_j = SD(X^{(j)})$ 
  		\STATE $Z^{(j)}:=\dfrac{X^{(j)}-\mu_{j}}{\sigma_{j}}$ \COMMENT {Z is the normalization of X }
  		\ENDFOR
  		
  		\COMMENT {Formation of distance matrix $D_{n\times n}$} 
  		
  		\FOR{$i=1$ to $n$}
  		\FOR{$j=1$ to $n$}
  		\STATE	$D(i,j)=||Z_i-Z_j||$ \COMMENT{$Z_i$ is $i^{th}$ row of matrix $Z$}
  		\ENDFOR
  		\ENDFOR
  		

  		\COMMENT {Formation of Affinity matrix $A_{n\times n}$ }
  		\STATE $\sigma_D = SD(D)$
  		\FOR{$i=1$ to $n$}
  		\FOR{$j=1$ to $n$}
  		\STATE	$ A(i,j)=exp(\dfrac{-D(i,j)^2}{2*\sigma_D})$   
  		\ENDFOR
  		\ENDFOR

  		\COMMENT{Take histogram $H_{10\times 1}$ of affinity values in 10 bins} 
  		
  		\FOR{$i=1$ to $n$}
  		\FOR{$j=1$ to $n$}
  		\STATE $H(\lceil A(i,j)*10 \rceil)=H(\lceil A(i,j)*10 \rceil)+1$
  		\ENDFOR
  		\ENDFOR

  		\COMMENT{Finding affinity threshold}
  		
  		
  		\STATE	$ k = \smash{\displaystyle\max_{ i }}\,(H(i+1)-H(i)), i \in \{1..9\}$\\
  		\STATE	$threshold=0.1(k-1)+0.05  $\\
  		
  		\COMMENT{Initialize vectors to zero. Cluster $C$, Size $S$ and Centroid $Cnt$}
  		\STATE$ C_{n \times 1} = \textbf{0}$ ; $ S_{n \times 1} = \textbf{0}$ ; $ Cnt_{n \times d} = \textbf{0}; k=0$ \\

  		\FOR{$i=1$ : $n$}

  		\IF{$C(i)==0$}

  		\STATE $k=k+1$
  		\STATE$C(i)=k$;	$S(k)=1$; $Cnt(k)=Z_i$
  		  		\COMMENT{If $i^{th}$ point does not belong any cluster then start forming a new ($k^{th}$) cluster with it.}
  		\ENDIF
  		
  		\FOR{$j=1$ : $n$}
  		\IF{$C(j)==0$}
	  		\IF{$A(Cnt(k),Z_j)> threshold$}
		  		\STATE$Cnt(k)=addPoint(Cnt(k),j)$; $C(j)=k$
		  		\COMMENT{Check unclustered points to include in current cluster}
		  	\ENDIF
		
  		\ELSIF{$D(Cnt(k),j)<D(Cnt(C(j)),j)$}
	  		\STATE $Cnt=removePoint(Cnt,j)$
	  		\STATE $Cnt(k)=addPoint(Cnt(k),j); $$C(j)=k$
	  		\COMMENT{Check clustered points to shift into current cluster}
  		\ENDIF
  		\ENDFOR
  		\ENDFOR \\
  		
  		\COMMENT{Remove outliers} 
  		\FOR{$i=1$ : $n$}	
  		\IF{$S(C(i))==1$}
  		\PRINT $X_i$ as outlier.			
  		\ENDIF
  		\ENDFOR
  		\STATE mergeClusters(Z,C,S)
  		
  	\end{algorithmic}
  \end{algorithm}

    \begin{algorithm}
    	\caption{$addPoint$}
    	\label{alg2}
    	\begin{algorithmic}[1]
    		  		\renewcommand{\algorithmicrequire}{\textbf{Input:}}
    		  		\renewcommand{\algorithmicensure}{\textbf{Output:}}
     		  		\REQUIRE $Cnt(k),Z_j,S$ \COMMENT{centroid of $k^{th}$ cluster $Cnt(k)$, point $j$, size vector S}
    		  		\ENSURE $Cnt(k)$ \COMMENT{centroid after adding $Z_j$}\\
  
		\STATE   $Cnt(k)=\frac{S(k)*Cnt(k) + Z_j}{S(k)+1} $
  		\STATE   $S(k)=S(k)+1;$ \COMMENT{size of $k^{th}$ cluster increased by 1}
  		\RETURN $Cnt(k)$
  		
  		\end{algorithmic}
  \end{algorithm}

    \begin{algorithm}
    	\caption{$removePoint$}
    	\label{alg3}
    	\begin{algorithmic}[1]
    		\renewcommand{\algorithmicrequire}{\textbf{Input:}}
    		\renewcommand{\algorithmicensure}{\textbf{Output:}}
    		\REQUIRE $Cnt,Z_j,S$ 
    		\COMMENT {Centroid vector $Cnt$, point $j$ to be removed, size vector S}
    		\ENSURE $Cnt$ 
    		\COMMENT{centroid after removing $j$}
    		
			\STATE	$ k'=C(j)$ \COMMENT{Point $j$ belongs to cluster $k'$}
    		\STATE  $ Cnt(k')=\frac{S(k')*Cnt(k') - Z_j}{S(k')-1}$
    		\STATE  $S(k')=S(k')-1;$ \COMMENT{size of $k^{th}$ cluster decreased by 1}
    		\RETURN $Cnt$
    	\end{algorithmic}
    \end{algorithm}

\subsection{Merging clusters}

	Second part of the algorithm merges the clusters if necessary. The problem with SSW is that it always increases when we  decrease number of clusters, thus merging two close clusters never reduces the cost. We define the cost as a slight modification to SSW: 
	\begin{equation}
W(C) = \sum\limits_{j=1}^p \frac{1}{s_j}\sum\limits_{\forall i|C(i)=j} (Z_i-Cnt(j))^2
	\end{equation}
	
	Unlike SSW, the proposed cost function W would decrease when closely lying clusters are merged. The decrease in cost would signify that data is favorable towards merge. For clusters, which are well separated, merging would increase the cost, prompting the merge to be discarded.

	Let $\{s_1,s_2,....s_p\}$ be the size of the p clusters in descending order, as detected in $findClusters$. We observe, in most datasets that supports merge, big size clusters are few in number while small size clusters are frequent. In estimating number of clusters for such dataset, the more skewed the distribution of sizes, the less number of clusters become appropriate to represent the data. Hence we prioritize the big size clusters as the representative. To get the number of clusters after merge, we take minimum k, such that, total differences between sizes of the bigger clusters  weighted by respective cluster size from $s_k$, exceeds that of the smaller clusters, i.e,
	\begin{equation}
		\min\limits_{k}\{ \sum\limits_{i=1}^{k-1}(s_i-s_k)s_i>\sum\limits_{j=k+1}^p(s_k-s_j)s_j\}
	\end{equation}
	where k gives the number of cluster in the dataset.
	\begin{figure}[!t]
		\includegraphics[scale=0.5]{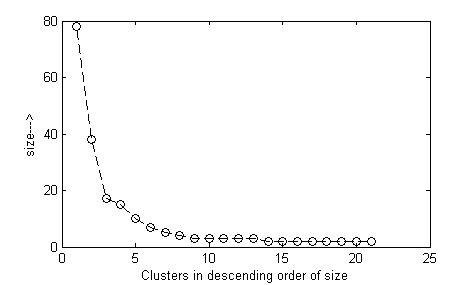}
		\caption{Initial cluster sizes of thyroid dataset}
		\label{fig2}
	\end{figure}

Fig. 2 shows the distribution of cluster sizes as found in the thyroid dataset\cite{UCI}. The number of clusters satisfying the equation 8, is k=2, coinciding the ground truth number of clusters of the dataset.

We then form a $p\times p$ matrix to get distance between all clusters using centroid distance that could be replaced by any other cluster distance measure like single, average or complete linkage as per the requirement\cite{slink,avg}. We merge two clusters which are closest in each step, $p-k$ such merge would produce the required k clusters, given by vector C'. 

Finally we compare cost after merge $W(C')$ with that of before merge $W(C)$, to ensure that merge was necessary. The merge is discarded in favor of initially detected clusters, if the cost increases after merge. Algorithm 4, $mergeClusters$ lists the pseudo code of this process. 

Fig. 3 shows datasets clustered by the algorithm before and after the merge. Dataset S2 is detected with $49$ clusters with $W(C)=7.77$ shown in 3(a), when merged it produces $15$ cluster with $W(C')=2.95$, indicating acceptance of merge in 3(b). In case of dataset R15, before merge there were $15$ clusters with $W(C)=1.84$, after merge it gives $6$ clusters with $W(C')=2.58$ as in 3(d), discarding the merge in favor of initially detected clusters in 3(c).

Fig. 4 shows in first two dimensions the d8c8N data set clustered by the algorithm, here outliers as desired are not included in any cluster.

\begin{figure}[!t]
			\includegraphics[scale=0.3]{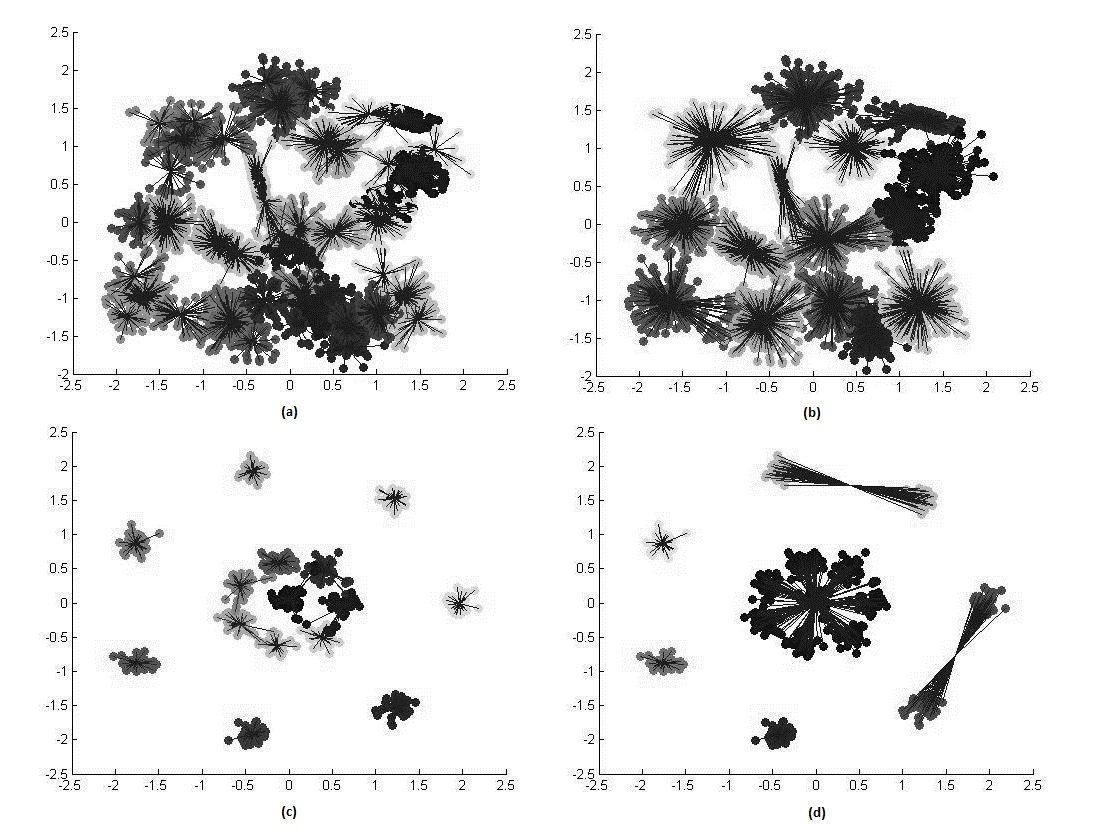}
			\caption{ S2 and R15 before and after the merge}
			\label{mrg}
\end{figure}
\begin{figure}[!t]
	\includegraphics[scale=0.29]{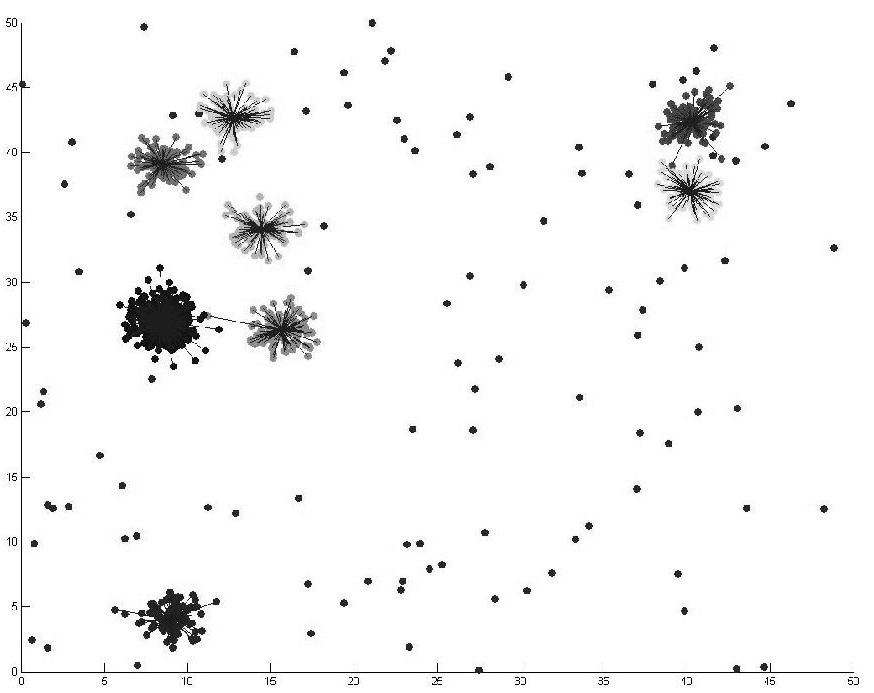}
	\caption{Clusters and Outliers detected in d8c8N}
	\label{outlier}
\end{figure}

 \begin{algorithm}
	\caption{$mergeClusters$}
	\label{alg4}
	\begin{algorithmic}[1]
			  \renewcommand{\algorithmicrequire}{\textbf{Input:}}
			  \renewcommand{\algorithmicensure}{\textbf{Output:}}
			  \REQUIRE $Z_{n\times d}$, $C_{n\times 1}, S$ \COMMENT{Normalized input matrix Z, cluster vector C, $S(i)$ denotes the size of $i^{th}$ cluster}
			  \ENSURE $C_{n\times 1}$ 
		\COMMENT{ Let p be remaining number of clusters after removing outliers}

		\COMMENT{Find k which satisfies equation (8)}
		
		\FOR{$k=2:p$}
		\STATE t1=0; \COMMENT{temp variable}
		\FOR{$i=1:k$}
		\STATE $ t1=t1+(S(i)-S(k))*S(i)$
		\ENDFOR
		
		\STATE t2=0; \COMMENT{temp variable}
		\FOR{$i=k+1:p$}
		\STATE $t2=t2+(S(k)-S(i))*S(i)$
		\ENDFOR
		\IF{$t1>t2$}
		\PRINT k; \COMMENT{number of clusters satisfying eqn. 8}
		\STATE break;
		\ENDIF

		\ENDFOR

		\STATE C'=C
		
		\FOR{$i=1:p-k$}
		
		\STATE	Merge two clusters that are closest; Update C'.
		
		\ENDFOR	\\
		
		\STATE 	calculate W(C) and W(C'); \COMMENT{ where $W(C) = \sum\limits_{j=1}^p \frac{1}{S(j)}\sum\limits_{C(i)=j} (Z_i-Cnt(j))^2$}
		\IF{$W(C')>W(C)$} 
		\STATE Discard merge.
		\PRINT number of clusters p
		\RETURN C
		
		\ELSE
		
		\PRINT number of clusters k 
		\RETURN C'
		
		\ENDIF\\

	\end{algorithmic}	
	\end{algorithm}

\subsection{Complexity analysis}
	Line 5, 11 and 16, $findClusters$ considers the all pair of distances with time and space complexity of $O(n^2)$. The loop started at line 24 also has same complexity as addition or removal of point from a cluster is performed incrementally in constant time by the functions $addPoint$ and $removePoint$. Thus, the worst case complexity of $findClusters$ is $O(n^2)$.  
	
	$mergeClusters$ estimates the number of clusters in the loop of line 1 that has time complexity  of $O(p^2)$, where $p (<n)$ is the number of clusters detected. As merging two clusters is an $O(n)$ operation, loop at line 16 has $O(pn)$ time complexity. It would required $O(p^2)$ space to store the distance between all cluster centroids. 
	
	Hence the proposed method which invokes algorithm 1 and 2 sequentially has a time and space complexity of the order of $O(n^2)$.

\section{Results}
Experiments are performed to understand the relative performance of the proposed algorithm with respect to the existing algorithms. Two aspects of the algorithm namely, estimation of number of clusters and quality of the obtained clusters are used as performance metric.

\subsection{Datasets}
Datasets used for cluster analysis generally have known number of clusters and are called ground truth datasets. Often they also have additional column telling which point belongs to which cluster as the output vector $C$ of our algorithm does. Evidently these meta information is used to verify performance of the algorithm and not given as the input to the algorithm. 
Table 1 shows real world and synthetic datasets used. They are introduced by respective researchers mentioned in the reference column and are available online\cite{datasets}. Dimension of the datasets varies from 2 to 1024, while maximum number of points in a dataset is 5401.  Last five datasets have density varying clusters, with 10 percent noise/outliers w.r.t original number of points, these are the most difficult dataset that are used in the comparative study of cluster validity indexes\cite{cvi_comp}. Their naming is done by the authors such that 'd' and 'c' denotes the dimension and number of clusters respectively, and 'N' stands for noise.

By the nature of the centroid distance function used, the algorithm would not work very well for shape based datasets, thus all synthetic data sets taken, are convex.
\begin{table}[!t]
	\begin{center}
		\begin{tabular}{ |c|c|c|c|c|c| } 
			\hline
			Dataset & Dimension & Points & Clusters & Type & Refernce\\ 
			\hline
			Iris & 4 &150 & 3 & Real & \cite{UCI} \\ 
			Thyroid & 5 & 215& 2 & " & " \\
			Car & 6 & 1728& 4 & " & " \\ 
			Breast & 9 & 699& 2 & " & " \\ 
			WDBC & 30 & 569 & 2 & " & " \\ 
			S1 & 2 & 5000 & 15 & Syn & \cite{S-set}\\ 
			S2 & 2 & 5000 & 15 & " & " \\ 
			S3 & 2 & 5000 & 15 & " & " \\ 
			S4 & 2 & 5000 & 15 & " & " \\
			D31 & 2 & 3100 & 31 & " & \cite{R15} \\
			R15 & 2 & 600 & 15 & " & " \\
			Uniform & 2 & 2000 & 1 & " & \cite{datasets} \\
			Diagonal & 2& 200 & 2 & " & " \\
			Dim2 & 2 & 1351 & 9 & " & " \\
			Dim4 & 2 & 2701 & 9 & " & " \\
			Dim8 & 2 & 5401 & 9 & " & " \\
			Dim032 & 32 & 1024 & 16 & " & \cite{dim-set}\\
			Dim064 & 64 & 1024 & 16 & " & "\\
			Dim128 & 128 & 1024 & 16 & " & "\\
			Dim256 & 256 & 1024 & 16 & " & "\\
			Dim512 & 512 & 1024 & 16 & " & "\\		
			Dim1024 & 1024 & 1024 & 16 & " & "\\
			d8c8N & 8 & 880 & 8 & " & \cite{cvi_comp}\\
			d8c4N & 8 & 880 & 4 & " & "\\
			d8c2N & 8 & 880 & 2 & " & "\\
			d4c4N & 4 & 770 & 4 & " & "\\
			d4c2N & 4 & 550 & 2 & " & "\\
			\hline
		\end{tabular}
	\end{center}
	\caption{Datasets}
	\label{tabl1}
\end{table}

\subsection{Estimation of cluster number}

 Correctness of cluster number estimated by the algorithm is compared with other available methods. For the comparison we have taken nine indexes which have best performance in different datasets\cite{cvi_comp,bin_comp}. Standard implementation available for the indexes\cite{R} are used  with k-means++ as clustering algorithm. Cluster number which is the most reported in hundred test runs are taken as reported value, in contrast the  proposed algorithm requires single run, as the output does not change if the order of the input data points remains the same. An exact match of reported value with the known number of clusters in the ground truth dataset is considered as success, we define accuracy as the percentage of datasets for which an algorithm correctly predicts the number of clusters.

\begin{table*}[]
	\begin{center}
	\begin{tabular}{|c|c|c|c|c|c|c|c|c|c|c|c|}
		\hline
		Dataset  &\textbf{Given}& CH \cite{CH}  & DB\cite{db}   & Silhouette\cite{silhouette} & Gap\cite{gap}  & Dunn \cite{dunn} & KL\cite{kl} & RL\cite{RL} & Scott\cite{scott} & Cindex\cite{cindex} & \textbf{Proposed} \\ \hline
		Iris     & 3     &\textbf{3}& 2     & 2          & 2  & 4   & 18    & 2         &\textbf{3}& 2      & 4   \\
		Thyroid  & 2     & 3     & 3     & 3          & 16 & 2   & 3     & 3         & 7     & 6      &\textbf{2}\\
		Car      & 4     & 3     & 18    & 3          & 20 & 27   & 3     & 5         &\textbf{4}& 14  & na    \\
		Breast   & 2     &\textbf{2}&\textbf{2}&\textbf{2}& 16 & 2 &\textbf{2}&\textbf{2}& 6      & 17  &\textbf{2}\\
		Wdbc     & 2     & 7     &\textbf{2}&\textbf{2}& 7   & 2  & 19    &\textbf{2}  & 4      & 12  &\textbf{2}\\
		S1       & 15    & 16    &\textbf{15}&\textbf{15}& 16 & 4 & 9     & 3          & 8      & 17  & 11       \\
		S2       & 15    & 16    & 12    & 12         &\textbf{15}& 13 & 12 & 3          & 4      & 20  &\textbf{15}\\
		S3       & 15    & 19    & 10    & \textbf{15}& 17  & 14  & 10     & 2          & 4      & 17  &\textbf{15}\\
		S4       & 15    & 16    & 13    & 16         & \textbf{15}& 13 & 3 & 3          & 3      & 16  & 9        \\
		D31      & 31    & 40    & 19    & 36         & 20  & 14  & 7     & 3           & 5     &\textbf{31} & 14       \\
		R15      & 15    & 19    & 6     & 16         & 16 & 6   & 19    & 3           & 11    & 16     &\textbf{15}\\
		Uniform  & 1     & na 	 & na   & 20         & \textbf{1} & na & na  & na    & na   & na    &\textbf{1}\\
		Diagonal & 2     & 6     &\textbf{2}&\textbf{2} &\textbf{2}&\textbf{2}&\textbf{2}&\textbf{2}& 4  & 3    &\textbf{2}\\
		Dim2     & 9     & 16    & 6     & 11         & na  & 5 & 16    & 5         & 15    & 20     &\textbf{9}\\
		Dim4     & 9     & 15    & 6     & 6          & 11   & 5 & 15    & 3         & 5     & 20     &\textbf{9}\\
		Dim8     & 9     & 15    & 2     & 12         & 14   & 2 & 15    & 4         & 4     & 20     &\textbf{9}\\
		Dim32    & 16    & 17    & 15    & 15         & 17   & 5 & 15    & 5         & 3     & 17     &\textbf{16}\\
		Dim64    & 16    & 19    & 20    & 19         & 20   & 4 & 19    & 5         & 4     & 19     &\textbf{16}\\
		Dim128   & 16    & 18    & 18    & 18         & na  & 6 & 18    & 7         & na & 18     &\textbf{16}\\
		Dim256   & 16    & 17    & 18    & 17         & na  & 11 & 17   & 11        & na & 17     &\textbf{16}\\
		Dim512   & 16    & 17    & 20    & 17         & na  &  5 & 17   & 8         & na & 17     &\textbf{16}\\
		Dim1024  & 16    &\textbf{16} & 20&\textbf{16}& na  & 13 & 20   & 14   & na & \textbf{16} &\textbf{16}\\
		d8c8N	 & 8	 & 16 	 & 2	 & 9 		  &	20	 &	9 & 10	 & 3 	& 3   & 17 			& \textbf{8}\\
		d8c4N	 & 4 	 &	6 	 & 3	 & 8		  & 18	 &	5 &	6	 & 3   	& 3	  & 10			& \textbf{4}\\
		d8c2N	 & \textbf{2}	 & \textbf{2} 	 & \textbf{2}	 & 3		  & 19 	 &	6 & 20   & \textbf{2} 	& 3   &	18 	& \textbf{2}	\\
		d4c4N	 & 4	 & \textbf{4}	 & \textbf{4}	 & \textbf{4}		  & 9    &	\textbf{4} &	13   & \textbf{4}	& \textbf{4}	  & 14	& 5\\
		d4c2N	 & 2	 & \textbf{2}	 & \textbf{2}	 & 3 		  & 6 	 &	\textbf{2} &	7	 & \textbf{2}	& 3	  & 20	& \textbf{2}	\\
		\hline
		\textbf{Match}	& Out of 27	& 6 & 7 & 7 & 4 & 6 & 2 & 6 & 3 & 2 & 21\\
		\hline
		
	\end{tabular}
\end{center}
	\label{tabl2}
	\caption{Relative performance to predict number of clusters}
	
\end{table*}
\begin{table*}
	\begin{center}
		\begin{tabular}{ |c|c|c|c|c|c|c|c|c|c| } 
			\hline
			Index& Method & R15 & s2 & s3 & dim32 & dim512 & d8c8N & d8c4N & d8c2N  \\
			\hline
			\multirow{6}{4em}{ARI} & Proposed & 0.922628 & 0.848518  & 0.599943 & \textbf{0.998958}  & \textbf{1} & \textbf{1}& \textbf{1}& \textbf{1} \\ 
			& k-means++ & 0.799796 & 0.836981 & 0.659792 & 0.779291 & 0.738598  & 0.758787 & 0.858261 &	0.871248	  \\
			& Ward      & 0.981996	 & 0.90575 & \textbf{0.677113} & \textbf{1} & \textbf{1} & 0.941304 & 0.930893 & 0.927027 \\
			& Average    & \textbf{0.98926}	& \textbf{0.912602} & 0.595741 & \textbf{1} & \textbf{1} &  0.551328 & 0.297042 & 0.040588 \\
			& Single & 0.542457 & 0 & 0 &	\textbf{1} & \textbf{1} & 0.002464 &	0.005564 & 0.008102 \\
			& GM & 0.827572 &	0.72655	& 0.449699 & na & na & 0.690421 & 0.722716 & 0.93182\\
			
			\hline
			\multirow{6}{4em}{Jaccard} & Proposed & 0.865108 & 0.752304 & 0.4603  & \textbf{0.998047} & \textbf{1} & \textbf{1}& \textbf{1}& \textbf{1}  \\ 
			& k-means++ & 0.691081 & 0.740213 & 0.519859 & 0.642106 & 0.612403 & 0.681939 & 0.836349 & 0.901123 \\
			& Ward      &  0.966894	& 0.83831 & \textbf{0.53714} & \textbf{1} & \textbf{1} & 0.907208 & 0.91259 & 0.941397 \\
			& Average   & \textbf{0.980118}   & \textbf{0.849156} & 0.456899 & \textbf{1} & \textbf{1} & 0.480418 & 0.443552 &0.577538 \\
			& Single & 0.4126 & 0.066575 & 0.066553 & \textbf{1} & \textbf{1} & 0.165506 & 0.329498 & 0.571032\\
			& GM  & 0.718604	& 0.596033	& 0.362868	& na & na & 0.588554 & 0.708168 & 0.94504 \\
			
			\hline
			\multirow{6}{4em}{F1}& Proposed & 0.927676  & 0.858645 & 0.630418 & \textbf{0.999022} & \textbf{1} & \textbf{1}& \textbf{1}& \textbf{1} \\ 
			& k-means++ & 0.809742 & 0.848396 & 0.683411  & 0.776575 & 0.754036  & 0.801726 & 0.909299 & 0.94799 \\
			& Ward 	 & 0.983168 & 0.912044 & \textbf{0.698882} & \textbf{1} & \textbf{1} & 0.951347 & 0.954298 & 0.969814 \\
			& Average & \textbf{ 0.989959} & \textbf{0.918426} & 0.627221 & \textbf{1} & \textbf{1} & 0.64903& 0.614528 & 0.732202 \\
			& Single & 0.584171 &	0.124838 & 	0.1248 &	\textbf{1} &	\textbf{1}	& 0.284007 & 0.495673 &	0.726951 \\
			& GM & 0.836264 & 0.743898 & 0.531798 & na & na & 0.740993 & 0.829155 & 0.971743 \\
			
			\hline
			
		\end{tabular}
	\end{center}
	\label{tabl3}
	\caption{Quality of clusters}
\end{table*}
Table 2 shows the performance of the algorithms in this respect. Out of 27 dataset taken the proposed algorithm has correctly estimated the number of clusters in 21 dataset.  We show  number of clusters for dataset car as not available(na) because clusters detected are more than $\sqrt{n}$. The algorithm is not able to find suitable clusters in the dataset as some attributes are categorical. Never the less it can be seen that the algorithm drastically outperforms the other methods, with an accuracy of 77.7\% as shown in Fig. 5, while among the other indexes taken Silhoutte and Davis-Bouldin given accuracy of 26.9\%. We attribute the success of the proposed algorithm to the high dimensional data and noise handling capability.
  	\begin{figure}[!t]
  		\includegraphics[scale=0.4]{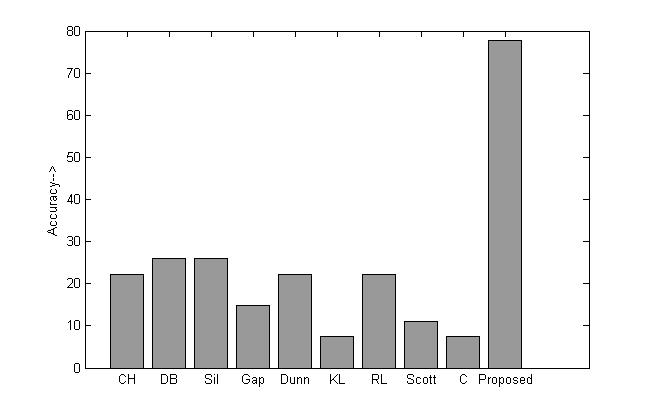}
  		\caption{Accuracy of cluster number estimation}
  		\label{fig4}
  	\end{figure}

\subsection{Quality of clusters}

Now we compare the quality of clusters obtained from the proposed algorithm with that of the existing algorithms. Most of the existing algorithms requires some parameter, in order to compare we have taken algorithms for which number of cluster could be supplied as parameter. Datasets for which correct cluster number are predicted by the proposed algorithm in first step, are compared with other clustering algorithms by supplying them the ground truth number of clusters as parameter.
We made comparison with kmeans++\cite{kmeans++} from partitioning method, single, average and ward from hierarchical method, and Gaussian mixture model with EM algorithm.

External evaluation indexes evaluates quality of clustering by comparing partition obtained from a clustering with actual partitions given in ground truth data. Adjusted Rand Index\cite{ARI}, Jaccard index and  F1 measure are standard such indices used for this purpose, all of them increase to represent better a match,the maximum value being unity.

Table 3 shows the quality of the clusters obtained in such format. In the noised data the algorithm outperforms others, while in other data the performance is comparable.
Among the eight datasets 62.5\% time the proposed algorithm has outperformed others, followed by that of 50\% time in Average linkage and 37.5\% time in  Ward's method. The relative performance of the algorithms remains the same w.r.t.  the different external evaluation indices.

\subsection{Bin analysis}
Here we analyze performance of the algorithm by varying number of bins to take affinity histogram in method $findClusters$. We have used accuracy as defined in the section 4.2 to study optimal number of bins. The proposed algorithm is executed on the 27 datasets mentioned in Table 1 with number of bins varying from 2 to 30. Figure 6 shows result of that experiment showing accuracy is maximum when number of bins are 10. The performance drops with less number of bins as variation in count with affinity is not properly captured, on the other hand with to many bins the difference subjects to local fluctuations in counts. The proposed algorithm takes number of bins as 10 to report performance.
  	\begin{figure}[!t]
		\includegraphics[scale=0.39]{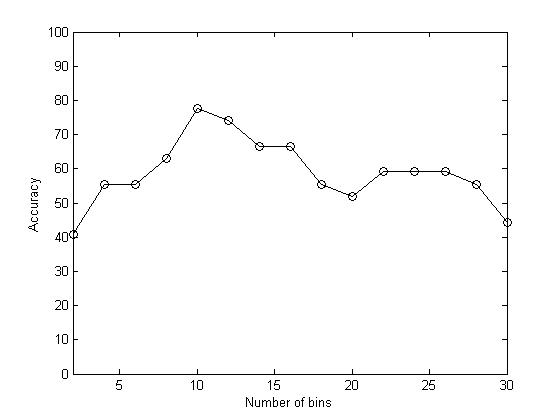}
  		\caption{Accuracy versus number of bins}
  		\label{fig5}
  	\end{figure}

\section{Conclusion}
We have proposed a parameter free clustering algorithm which gives promising results in convex datasets. The algorithm shows robustness to outliers, density variations and large dimension input data. The algorithm can be used dedicatedly to predict number of clusters where it outperformed existing indexes of estimating cluster numbers. The algorithm also works in a deterministic way when the order of input data is not changed, so that single run of the algorithm is sufficient. The future work might focus on reducing the complexity of the algorithm that is currently $O(n^2)$ and catering shape based non-convex data.

\section{Acknowledgement}
We profusely thank Mr. Saumik Bhattacharya, Department of  Electrical Engineering, IIT Kanpur for his scrupulous review and valuable feedback on the paper.


%

%
%

\ifCLASSOPTIONcaptionsoff
  \newpage
\fi


%
\bibliography{Reference}
\bibliographystyle{IEEEtran}

%

\begin{IEEEbiographynophoto}{Bhaskar Mukhoty}
is an M.Tech student in department of Computer Science Engineering, IIITDM Jabalpur. He has done B.Tech in CSE from Guru Nanak Institute of Technology, Kolkata. His research areas are Machine learning  and Stochastic modeling of random processes.
\end{IEEEbiographynophoto}

\begin{IEEEbiographynophoto}{Ruchir Gupta}
is Asst. Professor in the department of Computer Science Engineering, IIITDM Jabalpur.
He has done B.Tech. from HBTI Kanpur. His Ph.D. is from IIT Kanpur. His research
interests include Overlaid multicasting, Peer-to-peer
networking and information diffusion in distributed
network.
\end{IEEEbiographynophoto}

\begin{IEEEbiographynophoto}{Yatindra Nath Singh}
	is a professor in Electrical
Engineering department at IIT Kanpur. He
has done his B. Tech. from REC Hamirpur. His
M. Tech and Ph.D. are from IIT Delhi. He is a
senior member of IEEE, Fellow of IETE and
senior member, ICEIT. He had been Chairman of
IEEE UP section. He has been given AICTE
Young Teacher Award in 2002-03. His research interests
include Optical networks - protection and
restoration Packet and circuit switching, Optical
Packet switching architectures , Overlaid multicasting, Technology development
for E-learning and E-education , Peer to peer networking ,
Optical Communication systems and theory.
\end{IEEEbiographynophoto}





\end{document}